\documentclass[sigconf]{acmart}
\AtBeginDocument{
  \providecommand\BibTeX{{
    \normalfont B\kern-0.5em{\scshape i\kern-0.25em b}\kern-0.8em\TeX}}}

\usepackage{bbm}
\usepackage{makecell}
\usepackage{multirow}

\usepackage[section]{placeins}
\usepackage{amssymb}
\usepackage{amsmath,amsfonts}
\usepackage{algorithmic}
\usepackage{graphicx}
\usepackage{textcomp}
\usepackage{xcolor}

\usepackage{url}
\usepackage{booktabs}
\usepackage[linesnumbered,ruled]{algorithm2e}
\usepackage[inkscapeformat=png]{svg}
\usepackage{soul}
\usepackage{enumitem}
\usepackage{xcolor}
\usepackage{multirow}
\usepackage{subfig}
\usepackage{colortbl}
\usepackage{amsmath}
\usepackage{balance}

\copyrightyear{2024} 
\acmYear{2024}
\setcopyright{acmlicensed}
\acmConference[KDD '24]{Proceedings of the 30th ACM SIGKDD Conference on Knowledge Discovery and Data Mining}{August 25--29, 2024}{Barcelona, Spain}
\acmBooktitle{Proceedings of the 30th ACM SIGKDD Conference on Knowledge Discovery and Data Mining (KDD '24), August 25--29, 2024, Barcelona, Spain} \acmDOI{10.1145/3637528.3671710}
\acmISBN{979-8-4007-0490-1/24/08}

\definecolor{lgray}{rgb}{0.9,0.9,0.9}

\begin{document}

\title{Graph Data Condensation via Self-expressive Graph Structure Reconstruction}

% \author{Anonymous Authers}
\author{Zhanyu Liu}
\authornote{Both authors contributed equally to this work.}
\affiliation{
    \institution{Shanghai Jiao Tong University}
    \city{Shanghai}
    \country{China}
}
\email{zhyliu00@sjtu.edu.cn}

\author{Chaolv Zeng}
\authornotemark[1]
\affiliation{
    \institution{Shanghai Jiao Tong University}
    \city{Shanghai}
    \country{China}
}
\email{zclzcl@sjtu.edu.cn}

\author{Guanjie Zheng}
\authornote{Corresponding Author}
\affiliation{
    \institution{Shanghai Jiao Tong University}
    \city{Shanghai}
    \country{China}
}
\email{gjzheng@sjtu.edu.cn}

\begin{abstract}

With the increasing demands of training graph neural networks (GNNs) on large-scale graphs, graph data condensation has emerged as a critical technique to relieve the storage and time costs during the training phase. 
It aims to condense the original large-scale graph to a much smaller synthetic graph while preserving the essential information necessary for efficiently training a downstream GNN. 
However, existing methods concentrate either on optimizing node features exclusively or endeavor to independently learn node features and the graph structure generator. 
They could not explicitly leverage the information of the original graph structure and failed to construct an interpretable graph structure for the synthetic dataset. 
To address these issues, we introduce a novel framework named \textbf{G}raph Data \textbf{C}ondensation via \textbf{S}elf-expressive Graph Structure \textbf{R}econstruction (\textbf{GCSR}).
Our method stands out by (1) explicitly incorporating the original graph structure into the condensing process and (2) capturing the nuanced interdependencies between the condensed nodes by reconstructing an interpretable self-expressive graph structure. 
Extensive experiments and comprehensive analysis validate the efficacy of the proposed method across diverse GNN models and datasets.
Our code is available at \url{https://github.com/zclzcl0223/GCSR}.
\end{abstract}



\begin{CCSXML}
<ccs2012>
   <concept>
       <concept_id>10002951.10003227.10003351</concept_id>
       <concept_desc>Information systems~Data mining</concept_desc>
       <concept_significance>500</concept_significance>
       </concept>
 </ccs2012>
\end{CCSXML}

\ccsdesc[500]{Information systems~Data mining}

\keywords{Graph Data Condensation; Graph Neural Network; Node Classification}

\maketitle

\section{Introduction}

\begin{figure}[htbp]
    \centering
  \includegraphics[width=1.0\linewidth]{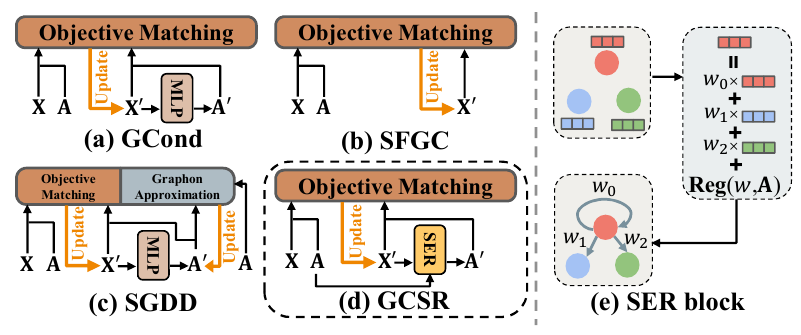}
  
  \caption{Learning pipelines of different graph condensation methods. $\textbf{X},\textbf{A}$ denotes the original full dataset and $\textbf{X}',\textbf{A}'$ denotes condensed dataset. \textit{SER} is our proposed self-expressive graph structure reconstruction module and \textit{Reg} is the regularization term for reconstruction.}
    \label{fig:comparison}
    \vspace{-.5cm}
\end{figure}

In recent years, the significant increase in the volume of graph data in different fields, such as social networks~\cite{naseem2021covidsenti,lamsal2021design}, recommendation systems~\cite{wu2020mind,eirinaki2018recommender}, traffic networks~\cite{liang2023cblab,liu2023cross,liu2024multi,liu2023fdti}, and knowledge graphs~\cite{zhang2020aser,wang2018acekg}, has presented challenges for practitioners and researchers. 
Consequently, storing, processing, analyzing, and transmitting large-scale data has become a burden in real-world applications.
Moreover, when considering deep learning tasks such as neural architecture search~\cite{ren2021comprehensive,liu2024dataset} and continual learning~\cite{han2020graph,li2017learning}, directly utilizing the full dataset for training a graph neural network (GNN) has the potential to result in poor efficiency.

To address this issue, data-centric graph size reduction methods, such as graph coarsening~\cite{cai2021graph,jin2020graph}, graph sparsification~\cite{batson2013spectral,chen2021unified}, coreset selection~\cite{li2017learning,welling2009herding}, and graph sampling~\cite{chiang2019cluster,zeng2019graphsaint}, aim to reduce the redundancy of the graph data and output a small dataset.
However, these methods rely on heuristics such as eigenvalues~\cite{batson2013spectral} and the accuracy loss~\cite{li2017learning}, which could lead to unsatisfactory generalization between different network architectures and suboptimal performance on downstream tasks such as node classification~\cite{jin2022condensing,yang2023does}.
In recent studies~\cite{jin2021graph, zheng2023structure, yang2023does, jin2022condensing, xu2023kernel}, the concept of \textit{Graph Data Condensation} has emerged, intending to train a compact synthetic dataset that exhibits comparable performance to the full dataset when utilized to train the same GNN model.
This approach achieves good results by matching surrogate objectives between the synthetic dataset and the original dataset.

However, existing graph data condensation methods fail to effectively and efficiently preserve the valuable information embedded within the original graph structure and capture the inter-node correlations within the condensed dataset.
Fig.~\ref{fig:comparison} illustrates the learning pipeline of various graph data condensation methods focused on node classification including GCond~\cite{jin2021graph}, SGDD~\cite{yang2023does}, SFGC~\cite{zheng2023structure}, and our proposed GCSR. 
GCond aims to model the graph structure by utilizing node features as input to a multilayer perceptron (MLP), while SFGC proposes a graph-free framework.
These two methods cannot build explicit and interpretable inter-node correlations, and ignore the rich information contained in the original graph structure.
SGDD employs the original graph structure to guide the generation of the synthetic graph structure through graphon approximation, but the application of bi-loop optimization may result in unstable performance and unsatisfactory efficiency.

To effectively and efficiently construct an explicit and interpretable graph structure for the synthetic condensed data, and integrate the information from the original graph structure, we propose a novel framework named \textbf{G}raph Data \textbf{C}ondensation via \textbf{S}elf-expressive Graph Structure \textbf{R}econstruction (\textbf{GCSR}).
Overall, GCSR contains three novel modules.
The first module is the Initialization Module, where we utilize the $k$-order node feature for node initialization and the probabilistic adjacency matrix derived from the original graph structure to initialize the regularization term used in the subsequent reconstruction process.
The second module is the Self-expressive Reconstruction Module.
Leveraging the self-expressiveness property of graph data~\cite{kang2022fine,xu2019scaled}, which indicates that nodes within the same feature subspace can represent each other, we reconstruct an explicit and interpretable graph structure using a closed-form expression.
The last module is the Update Module.
In this module, the node feature is updated through multi-step gradient matching with the training trajectories of the full dataset. 
The regularization term of the reconstruction is updated through bootstrapping, ensuring adaptability to the training dynamics and preventing overfitting of the learned graph structure to the small condensed dataset.
In general, our contributions can be summarized as follows:
\begin{itemize}[leftmargin=*]
    \item To the best of our knowledge, we are the first to attempt to construct the explicit and interpretable graph structure for the synthetic condensed data in the graph condensation task. The graph structure is constructed based on the self-expressive nature of graph data and effectively integrates the extensive information present in the original full graph.
    \item We propose a novel framework for the graph condensation task via self-expressive graph structure reconstruction, abbreviated as GCSR. 
    This framework encompasses three key modules, including initialization, self-expressive reconstruction, and update.
    These modules work in tandem to generate condensed data that effectively captures the essence of the original graph dataset.
    \item We conduct extensive experiments on five real-world graph datasets. 
    The results demonstrate that our proposed framework achieves superior performance and has many characteristics such as maintaining the inter-class similarity of the original graph. 
\end{itemize}

\section{Related Work}

\textbf{Graph Neural Networks.} Taking both node features and their structure information as the input, graph neural networks (GNNs)~\cite{kipf2016semi, defferrard2016convolutional, hamilton2017inductive, velickovic2017graph, wu2019simplifying, du2019graph, gasteiger2018predict, yang2022graph, xu2018powerful} have emerged as powerful tools in graph machine learning. 
The versatility of graph-structured data and the strong graph representation capabilities of GNNs have led to their widespread adoption in a range of real-world applications, including recommender systems~\cite{fan2019graph}, natural language processing~\cite{wu2023graph}, and computer vision~\cite{nazir2021survey}.

\vspace{2pt}\noindent \textbf{Dataset Condensation.} Dataset condensation (DC), also known as dataset distillation~\cite{wang2018dataset, zhao2020dataset, zhao2021dataset, nguyen2021dataset, cazenavette2022dataset, zhao2023dataset, liu2022dataset, sajedi2023datadam}, aims to generate a smaller synthetic dataset from a large training dataset, such that it can effectively substitute the original dataset for downstream training tasks. To obtain such a synthetic dataset, various works employ different optimization objectives. 
Some methods~\cite{nguyen2021dataset,wang2018dataset,nguyen2020dataset,liu2022dataset,deng2022remember,zhou2022dataset,loo2023dataset} learn the synthetic dataset by using a meta-learning style framework to optimize the performance on the real full data.
Some methods~\cite{lee2022dataseta,kim2022dataset,zhao2021dataset,zhao2020dataset,du2023minimizing,zhang2023accelerating,cazenavette2022dataset,cui2022scaling,li2022dataset} optimize the synthetic dataset by matching the gradients of the neural networks trained on the synthetic dataset and the original dataset respectively. 
Some methods~\cite{lee2022datasetb,wang2022cafe,zhao2023dataset,sajedi2023datadam} optimize the synthetic dataset by matching the distribution between the synthetic dataset and the original dataset. 
While initially applied to image data, DC has recently found wide application in graph-structured data for both node-level tasks~\cite{jin2021graph, zheng2023structure, yang2023does} and graph-level tasks~\cite{jin2022condensing, xu2023kernel}. GCond~\cite{jin2021graph} first tries to condense graph data via gradient matching and generate the adjacency matrix from the synthetic node features using a trained multilayer perception. 
SGDD~\cite{yang2023does} enhances GCond by broadcasting the original graph structure to the generation of the synthetic graph structure through laplacian energy distribution matching. 
SFGC~\cite{zheng2023structure} proposes to generate structure-free graphs by matching training trajectories and selecting optimal synthetic graphs using graph neural tangent kernel (GNTK)~\cite{du2019graph}. 
However, these methods could not effectively and efficiently capture an explicit and interpretable graph structure for the condensed graph and neglect the explicit information of the original graph structure, which results in suboptimal performance.

\vspace{2pt}\noindent \textbf{Self-expressiveness.} The self-expressiveness property indicates that each data sample can be represented by a linear combination of other data points~\cite{ma2020towards, lv2021pseudo}. 
Such a property is first explored in subspace segmentation~\cite{lu2012robust}, which aims to
segment data drawn from multiple linear subspaces as data in the same subspace could represent each other. 
Later, researchers begin to apply this property to graph learning~\cite{ pan2021multi, xu2019scaled}, aiming to learn a denoised and interpretable graph structure for downstream tasks such as graph clustering~\cite{kang2022fine, ma2020towards}.
They add different regularizations and constraints to ensure that the learned graph structure can fully reflect the connection relationships between nodes.
\section{Preliminary}
\noindent\textbf{Graph Condensation.}
A graph dataset $\mathcal{T}$ is denoted as $\mathcal{T}=\{\textbf{A}, \textbf{X}, \textbf{Y}\}$, where $\textbf{A}\in \mathbb{R} ^{N\times N}$ is the adjacency matrix, $\textbf{X}\in \mathbb{R} ^{N\times d}$ is the node feature, and $\textbf{Y}\in \mathbb{R} ^{N\times 1}$ is the label.
Generally, graph data condensation aims to learn a downsized synthetic graph $\mathcal{S}=\{\textbf{A}', \textbf{X}', \textbf{Y}'\}$ with $\textbf{A}'\in \mathbb{R} ^{N'\times N'}$, $\textbf{X}'\in \mathbb{R} ^{N'\times d}$, and $\textbf{Y}'\in \mathbb{R} ^{N'\times 1}$ from the original training graph $\mathcal{T}$ ($N' \ll N$). Both $\textbf{Y}_i$ and $\textbf{Y}'_j$ belong to the class set $\mathcal{C}=\{0,1,\cdots,C-1\}$.
Formally, graph condensation is defined by solving the problem as follows:
\begin{equation}
    \begin{aligned}
        &\min \mathcal{L}(\text{GNN}_{\theta_{\mathcal{S}}}(\textbf{A},\textbf{X}),\textbf{Y})\\
        s.t.\quad &\theta_{\mathcal{S}}=\underset {\theta} { \operatorname {arg\,min} }\,\mathcal{L}(\text{GNN}_{\theta}(\textbf{A}',\textbf{X}'),\textbf{Y}')\,.
    \end{aligned}
\end{equation} 
Here, $\text{GNN}_{\theta}$ represents the GNN model parameterized with $\theta$, $\theta_{\mathcal{S}}$ denotes the parameters of the model trained on the synthetic graph $\mathcal{S}=\{\textbf{A}',\textbf{X}',\textbf{Y}'\}$, and $\mathcal{L}$ is the error of node classification.
For the graph data condensation task, only the variables $\textbf{A}'$ and $\textbf{X}'$ are optimized, while the node labels $\textbf{Y}'$ are fixed, maintaining the same class distribution as the original labels $\textbf{Y}$. 
\section{Method}

\begin{figure*}[!th]
    \centering
    \includegraphics[width=0.92\linewidth]{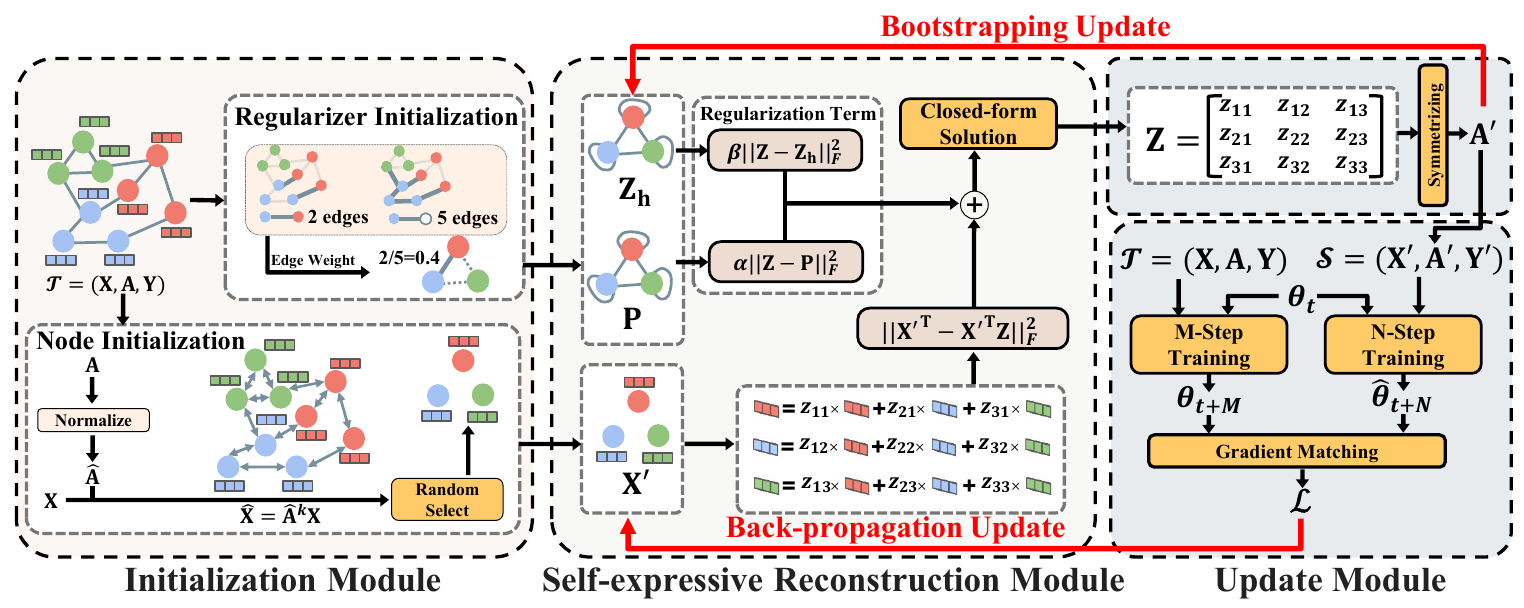}
    \caption{Overview of Graph Data Condensation via Self-expressive Graph Structure Reconstruction (GCSR).}
    \label{fig:model}
\end{figure*}

In this section, we present our proposed framework GCSR, which consists of three essential modules.
The diagram of the framework is depicted in Fig.~\ref{fig:model}, and the detailed pseudo code is shown in Alg.~\ref{alg1}.
The first module, termed the Initialization Module, initializes the node feature and graph regularizer for the downstream synthetic dataset.
Then, the Self-expressive Reconstruction Module exploits the inherent self-expressive property of the graph data and employs a closed-form solution to derive an interpretable graph structure that captures the self-representation relationships among the nodes.
Finally, the Update Module uses the derived adjacency matrix to update the graph regularizer in a bootstrapping style and the multi-step gradient matching loss to update the node feature respectively.

\subsection{Initialization}\label{sec:submp}

\subsubsection{Node Initialization}\hfill

\noindent\textit{\textbf{Goal:}}
The initial feature of the condensed dataset is crucial to the final performance.
Previous methods utilize random noise~\cite{wang2022cafe,deng2022remember} or random samples from the original dataset~\cite{zhao2023dataset,cazenavette2022dataset,jin2021graph,yang2023does}.
However, these methods overlook the valuable graph structure information during the node feature initialization stage.
In fact, in the field of graph learning, it has been observed that node features that incorporate message passing exhibit higher expressiveness and yield better performance in downstream tasks, such as node classification, due to the fusion of graph structure information~\cite{yang2022graph,wu2019simplifying}.
Building upon this insight, we propose an approach to initialize the node feature after the message passing process. 
By doing so, we ensure that the framework is initialized with better node features, leading to enhanced performance in subsequent tasks.

\vspace{2pt}\noindent\textit{\textbf{Message Passing Initialization:}}
Given the original graph dataset $\mathcal{T}=\{\textbf{A}, \textbf{X}, \textbf{Y}\}$, we first normalize it following previous work~\cite{kang2022fine,ma2020towards,kipf2016semi}.
The symmetrically normalized adjacency matrix can be defined as:
\begin{equation}
    \begin{aligned}
        \hat{\textbf{A}}=\tilde{\textbf{D}}^{-\frac{1}{2}}\tilde{\textbf{A}}\tilde{\textbf{D}}^{-\frac{1}{2}}\,,
    \end{aligned}
    \label{formula:2}
\end{equation} where $\tilde{\textbf{A}}=\textbf{A} + \textbf{I}$ is the adjacency matrix with self-connections added, $\tilde{\textbf{D}}_{ii}=\sum_j\tilde{\textbf{A}}_{ij}$ is the degree matrix, and $\textbf{I}$ is the identity matrix.
Under the framework of message passing mechanisms~\cite{gilmer2017neural,ma2020towards}, each node of $\mathcal{T}$ aggregates the information from its $k$-order neighborhood:
\begin{equation}
    \begin{aligned}
        \hat{\textbf{X}}=\hat{\textbf{A}}^k \textbf{X}\,,
    \end{aligned}
    \label{formula:3}
\end{equation} where $\hat{\textbf{A}}^k$ indicates the $k$-order adjacency matrix.

After Eq.~\ref{formula:2} and Eq.~\ref{formula:3}, we obtain a node representation $\hat{\textbf{X}}$ that aggregates neighborhood features as well as graph structure information. 
We then initialize the synthetic nodes $\textbf{X}'$ by randomly sampling the node feature from $\hat{\textbf{X}}$,
\begin{equation}
    \textbf{X}'=Random\_Sample(\hat{\textbf{X}})\,.
    \label{formula:4}
\end{equation}

Moreover, to make the synthetic dataset conform to the distribution of the original dataset and avoid creating an unbalanced dataset, we keep the class distribution the same as the original graph. Formally, the following equation holds for every node class, i.e., $c\in\{0,1,\cdots,C-1\}$.
\begin{equation}
    \sum_{i=1}^{N'}\frac{\mathbbm{1}(\textbf{Y}'_i==c)}{N'}=\sum_{k=1}^{N}\frac{\mathbbm{1}(\textbf{Y}_k==c)}{N}\,.
\end{equation}
Here, $\textbf{Y}'_i$ and $\textbf{Y}_k$ are the class of node in the synthetic dataset and original dataset respectively, and $\mathbbm{1}$ is the indicator function.

\subsubsection{Regularizer Initialization}\hfill

\noindent\textit{\textbf{Goal:}}
The downstream self-expressive reconstruction task needs a graph structure regularizer to avoid the solution falling into a trivial solution~\cite{xu2019scaled,kang2022fine}.
So, we aim to initialize a graph structure regularizer by incorporating the information of the original graph structure.
However, considering that the dimensions of adjacency matrices of the synthetic graph and the original graph are inconsistent, it is difficult to directly transfer the structure information of the original graph to the synthetic graph.
Consequently, we propose to generate a probabilistic adjacency matrix~\cite{franceschi2019learning, jin2022condensing} based on the information from the original adjacency matrix as the initial graph regularizer.
To guarantee the stability of the training process, we replace the learnable parameter and use the frequency of edge types as the probability instead, which reflects the correlation between classes.

\noindent\textit{\textbf{Graph Regularizer Initialization:}}
Given the original adjacency matrix $\textbf{A}_{ij}$, we aim to first calculate the frequency of edge types and construct a class-wise correlation matrix as follows:
\begin{equation}
    \overline{\textbf{A}}_{cc'}=\frac{\sum_{i,j=1}^N \mathbbm{1}(\textbf{Y}_i=c)\mathbbm{1}(\textbf{Y}_j=c')\textbf{A}_{ij}}{\sum_{i,j=1}^N \mathbbm{1}(\textbf{Y}_i=c)\textbf{A}_{ij}},
    \label{formula:5}
\end{equation} where $\overline{\textbf{A}}\in\mathbb{R}^{|\mathcal{C}|\times|\mathcal{C}|}$ represents the adjacency matrix generated from the original graph, $\textbf{Y}_i$ is the class of node $i$ of the original dataset, and $\mathbbm{1}$ is the indicator function. 
The value of $\overline{\textbf{A}}_{cc'}$ implies the frequency that an edge from a node in class $c$ to a node in class $c'$ exists. 
With $\overline{\textbf{A}}$, we can generate an adjacency matrix that leverages the structure information of the original graph:
\begin{equation}
    \begin{aligned}
    \textbf{P}_{ij}=\overline{\textbf{A}}_{\textbf{Y}'_i\textbf{Y}'_j}\,,
    \end{aligned}
    \label{formula:6}
\end{equation} where $\textbf{P} \in \mathbb{R}^{N'\times N'}$ and $\textbf{Y}'_i$ denotes the label of synthetic node $i$.
Consequently, $\textbf{P}$ captures the inter-class correlations and the correlations offer valuable information for the downstream self-expressive reconstruction.
By utilizing $\textbf{P}$ as the graph structure regularization term, the framework could avoid generating trivial solutions in the optimization process.

\begin{algorithm}[!t]
\caption{GCSR for Graph Data Condensation}
\renewcommand{\algorithmicrequire}{\textbf{Input:}}
\renewcommand{\algorithmicensure}{\textbf{Output:}}
\label{alg1}
\begin{algorithmic}[1]
\REQUIRE $\mathcal{T}=\{\textbf{A}, \textbf{X}, \textbf{Y}\}$: training data
\REQUIRE $\textbf{Y}'$: pre-defined synthetic labels
\REQUIRE $\{\theta\}$: set of expert GNN parameters trained on $\mathcal{T}$
\renewcommand{\algorithmicrequire}{\textbf{Require:}}
\REQUIRE Training epochs $K$, synthetic steps $N$, expert steps $M$, learning rate $\eta_1$, $\eta_2$, regularization weight $\alpha$, $\beta$, update rate $\tau$, $\gamma$
\STATE Initialize $\textbf{X}'$ according to Eq.~\ref{formula:3} and Eq.~\ref{formula:4}
\STATE Initialize $\textbf{P}$ according to Eq.~\ref{formula:5} and Eq.~\ref{formula:6}
\STATE Initialize $\textbf{Z}_h$ as identity matrix $\textbf{I}$
\FOR{$k=1\to K$} 
    \STATE Sample expert GNN parameters: $\theta_t\sim P_{\theta}$\;
    \STATE Initialize synthetic GNN parameters: $\widehat{\theta}_t = \theta_t$\;
    \STATE Compute $\textbf{A}'$ according to Eq.~\ref{formula:9} and Eq.~\ref{formula:10}\;
    \FOR{$i=1\to N$} 
        \STATE Update synthetic GNN w.r.t. classification loss: \\$\qquad\widehat{\theta}_{t+i}=\widehat{\theta}_{t+i-1}-\eta_1\nabla l(\mathcal{S},\widehat{\theta}_{t+i-1})$\;
    \ENDFOR
    \STATE Compute loss between synthetic and expert params:\\ $\qquad\mathcal{L}=||\widehat{\theta}_{t+N}-\theta_{t+M}||^2_F/{||\theta_{t+M}-\theta_{t}||^2_F}$\;
    \STATE Update $\textbf{X}'\leftarrow\textbf{X}'-\eta_2\nabla_{\textbf{X}'}\mathcal{L}(\mathcal{S},\mathcal{T})$\;
    \STATE Update $\textbf{P}\leftarrow\tau\textbf{P}+(1-\tau)\textbf{A}'$\;
    \STATE Update $\textbf{Z}_h\leftarrow\gamma\textbf{Z}_h+(1-\gamma)\textbf{A}'$\;
\ENDFOR

\tcp{Computing the final adjacency matrix $\mathbf{A}'$}
\STATE $\textbf{Z}=(\textbf{X}'\textbf{X}'^{T}+\alpha\textbf{I}+\beta\textbf{I})^{-1}(\textbf{X}'\textbf{X}'^{T}+\alpha\textbf{P}+\beta\textbf{Z}_h)$
\STATE $\textbf{A}' = (|\textbf{Z}|+|\textbf{Z}|^T)/2$
\ENSURE $\mathcal{S}=\{\textbf{A}', \textbf{X}', \textbf{Y}'\}$
\end{algorithmic}  
\end{algorithm}

\subsection{Self-expressive Reconstruction}
\subsubsection{Reconstruction}

In recent years, it has been witnessed the great success of graph structure learning \cite{pan2021multi, kang2022fine, ma2020towards} via self-expressive graph structure reconstruction.
Such a strategy exploits the self-expressiveness property of graph nodes to reconstruct the adjacency matrix in structure-free graphs. Mathematically, it can be formulated as the following expression:
\begin{equation}
    \min_\textbf{Z}||\textbf{X}'^{T}-\textbf{X}'^{T}\textbf{Z}||_F^2\,.
\end{equation}
Here, $\textbf{X}'\in\mathbb{R}^{N'\times d}$ is the node features, and $\textbf{Z}\in \mathbb{R}^{N'\times N'}$ is the self-expressive matrix that measures the similarity between nodes with interpretability.
However, solving such an equation could lead to trivial solutions such as identity matrix $\textbf{I}$.
Consequently, existing methods~\cite{elhamifar2013sparse,liu2010robust,liu2012robust,xu2019scaled,kang2022fine} impose various regularizations such as least-square and low rankness. For example, Least Square Regression (LSR) could be represented as:
\begin{equation}
    \begin{aligned}
        \min_\textbf{Z}||\textbf{X}'^{T}-\textbf{X}'^{T}\textbf{Z}||_F^2+\alpha ||\textbf{Z}||_F^2\,,
    \end{aligned}
    \label{formula:7}
\end{equation} where $\alpha>0$ is a trade-off parameter and $||\cdot||_F$ is Frobenius norm.
Nonetheless, the structure information of the original graph is only implicitly used when initializing $\textbf{X}'$ via message passing. 
To explicitly incorporate the original graph structure and the synthetic graph structure information to $\textbf{Z}$, we add a regularization term and propose our self-expressive reconstruction model:
\begin{equation}
    \begin{aligned}
        \min_\textbf{Z}||\textbf{X}'^{T}-\textbf{X}'^{T}\textbf{Z}||_F^2 +\alpha||\textbf{Z}-\textbf{P}||_F^2\,,
    \end{aligned}
\end{equation} where $\textbf{P}$ is the synthetic adjacency matrix generated from the original graph.
By incorporating $\textbf{P}$ into the regularization term, the similarity matrix could learn from the original graph structure.
However, since the synthetic node features $\textbf{X}'$ is updated in each epoch and $\textbf{X}'$ has a significant impact on the optimizing object, the learned similarity matrix $\textbf{Z}$ could experience drastic changes, resulting in unstable performance.
To address this issue and mitigate the fluctuations, we propose to introduce the historical similarity matrix $\textbf{Z}_h$ into the regularization term.
By including $\textbf{Z}_h$, which captures the past information of the similarity matrix, we ensure a more stable learning process and enhance the overall performance of the model.
\begin{equation}
        \min_\textbf{Z}||\textbf{X}'^{T}-\textbf{X}'^{T}\textbf{Z}||_F^2 +\alpha||\textbf{Z}-\textbf{P}||_F^2+\beta||\textbf{Z}-\textbf{Z}_h||_F^2\,.
        \label{formula:8}
\end{equation}
Here, $\textbf{Z}_h$ is initialized as identity matrix $\textbf{I}$ and is updated through the training process of the graph condensation task.

\subsubsection{Closed-form Solution}

Directly optimizing Eq.~\ref{formula:8} is time-consuming.
However, Eq.~\ref{formula:8} can be easily solved by setting its first-order derivative w.r.t. $\mathbf{Z}$ to zero:
\begin{equation}
    -2\mathbf{X}'(\mathbf{X}'^T-\mathbf{X}'^T\mathbf{Z})+2\alpha(\mathbf{Z}-\mathbf{P})+2\beta(\mathbf{Z}-\mathbf{Z}_h)=0\,.
    \label{eq:1}
\end{equation}
Eq.~\ref{eq:1} could be reduced to:
\begin{equation}
    \begin{aligned}
        \mathbf{Z}=(\mathbf{X}'\mathbf{X}'^{T}+\alpha\mathbf{I}+\beta\mathbf{I})^{-1}(\mathbf{X}'\mathbf{X}'^{T}+\alpha\mathbf{P}+\beta\mathbf{Z}_h)\,.
    \end{aligned}
    \label{formula:9}
\end{equation}

Here, $\mathbf{Z}$ not only takes the correlations between nodes into account (i.e., $\mathbf{X}'\mathbf{X}'^{T}$), but also retains the structure information of the original graph (i.e., $\mathbf{P}$). 
Furthermore, the learned similarity matrix $\mathbf{Z}$ possesses interpretability as it explicitly indicates the weight to which the node feature of one node is represented by the other nodes in the synthetic graph dataset.
This interpretability allows for a clear understanding of dependencies among the nodes of the synthetic graph dataset, leading to enhanced transparency in the proposed framework and a comprehensive comprehension of the graph data condensation task for the first time.

Considering the symmetry and non-negativity of the real-world adjacency matrix, we symmetrize $\mathbf{Z}$~\cite{kang2022fine,zhang2019attributed} and calculate the final synthetic adjacency matrix as follows:
\begin{equation}
    \begin{aligned}
        \mathbf{A}'=\frac{1}{2}(|\mathbf{Z}|+|\mathbf{Z}|^T)\,,
    \end{aligned}
    \label{formula:10}
\end{equation} where $|\cdot|$ is the elementwise absolute operation. $\mathbf{A}'\in\mathbb{R}^{N'\times N'}$ is the synthetic adjacency matrix.

\subsubsection{Complexity Analysis}
For Eq.~\ref{formula:9}, the time complexity of computing $\mathbf{X}'\mathbf{X}'^{T}$ is $O(N'^2d)$. 
The time complexity of computing $(\mathbf{X}'\mathbf{X}'^{T}+\alpha\mathbf{I}+\beta\mathbf{I})^{-1}$ is $O(N'^3)$ due to the matrix inverse operation. 
The time complexity of matrix multiplication between the two main terms is $O(N'^3)$.
Consequently, the overall time complexity for calculating Eq.~\ref{formula:9} is $O(N'^3+N'^2d)$. 
In practice, since the matrix multiplication could be highly parallelized, the main time-consuming part of Eq.~\ref{formula:9} is the matrix inverse operation with a complexity of $O(N'^3)$.
It is worth noting that this time complexity is acceptable since the condensed dataset is small, with $N'\approx 100$.

\subsection{Update}\label{sec:subupdate}

\subsubsection{Node Feature Update}

In order to update and refine the synthetic data $\mathcal{S}$ to ensure its performance similarity to the original data $\mathcal{T}$ when training downstream models, it is essential to enable the synthetic data $\mathcal{S}$ to learn the training dynamics observed in models trained with the real data $\mathcal{T}$.
This learning process facilitates the alignment of the synthetic data with the underlying patterns and characteristics present in the original data, thereby enhancing its suitability for downstream tasks such as node classification.
Here, we adopt multi-step gradient matching~\cite{cazenavette2022dataset,zheng2023structure}, which essentially matches the multi-step gradient of a model trained by the synthetic data $\mathcal{S}$ and the real data $\mathcal{T}$.
Formally, the matching objective could be formulated as follows:
\begin{equation}
\begin{gathered}
\begin{aligned}
  D(\mathcal{S},\mathcal{T};\theta_t)  & =   \cfrac{||(\widehat{\theta}_{t+N}-\theta_t) - (\theta_{t+M}-\theta_t)||_F^2}{||{\theta}_{t+M}-\theta_{t}||_F^2}\\
   & =  \cfrac{||\widehat{\theta}_{t+N}-{\theta}_{t+M}||_F^2}{||\theta_{t+M}-{\theta}_{t}||_F^2}\,,
\end{aligned}
\end{gathered}
\end{equation}
where $\theta_t$ is the model parameter sampled from the training trajectory of $\mathcal{T}$.
Based on $\theta_t$, we optimized it with dataset $\mathcal{S}$ for $N$ epochs to generate the trained parameter $\widehat{\theta}_{t+N}$ and optimized it with dataset $\mathcal{T}$ for $M$ epochs to generate ${\theta}_{t+M}$.

Subsequently, the matching loss is the expectation of the $D(\mathcal{S},\mathcal{T};\theta_t)$ w.r.t. $\theta_t$, which could be formulated as follows:
\begin{equation}
    \mathcal{L}(\mathcal{S},\mathcal{T})=\text{E}_{\theta_t\sim P_{\theta}}[D(\mathcal{S},\mathcal{T};\theta_t)]\,,
\end{equation}
where $P_{\theta}$ is the distribution of parameter trajectories trained by real dataset $\mathcal{T}$. 
By optimizing $\mathcal{L}(\mathcal{S},\mathcal{T})$, the synthetic graph dataset $\mathcal{S}$ would be guided by the training trajectories of the original graph $\mathcal{T}$.
Consequently, the model trained by $\mathcal{S}$ could have similar parameters to the model trained by $\mathcal{T}$ and the performance on downstream tasks is expected to be comparable.

\subsubsection{Graph Regularizer Update}

The graph regularizer $\mathbf{P}$ plays a significant role in determining the resulting graph structure $\mathbf{A}'$. 
However, maintaining $\mathbf{P}$ unchanged throughout the condensing process can lead to undesirable outcomes, such as the synthetic graph $\mathbf{A}'$ overfitting to the specific characteristics of $\mathbf{P}$. 
Furthermore, real-world graphs are commonly subject to noise and incompleteness~\cite{franceschi2019learning, liu2022towards}, and these imperfections could be inherited by the graph regularizer $\mathbf{P}$.
To address these issues, we update $\mathbf{P}$ with bootstrapping algorithms~\cite{liu2022towards,grill2020bootstrap} as follows:
\begin{equation}
    \begin{aligned}
        \mathbf{P}&\leftarrow\tau\mathbf{P}+(1-\tau)\mathbf{A}'\,.
    \end{aligned}
    \label{formula:14}
\end{equation}
Similarly, we update the historical similarity matrix $\mathbf{Z}_h$ as follows:
\begin{equation}
    \begin{aligned}
        \mathbf{Z}_h&\leftarrow\gamma\mathbf{Z}_h+(1-\gamma)\mathbf{A}'\,,
    \end{aligned}
    \label{formula:gamma}
\end{equation}
where $\tau\in[0,1]$ and $\gamma\in[0,1]$. $\tau\ge0.9$ provides a very slow update to preserve as much of the original structure information as possible. $\gamma\ge0.5$ provides a faster update to ensure that the newest structure information can be retained and benefit the downstream processes.

\section{Experiment}
\subsection{Experimental Settings}

\textit{\textbf{Datasets:}} In line with previous research~\cite{jin2021graph,zheng2023structure,yang2023does}, we evaluate our proposed framework on node classification task. 
We use three transductive datasets: Citeseer~\cite{kipf2016semi}, Cora~\cite{kipf2016semi}, and Ogbn-arxiv~\cite{hu2020open}, as well as two inductive datasets: Flickr~\cite{zeng2019graphsaint} and Reddit~\cite{hamilton2017inductive}.
The details of the original dataset statistics are shown in Table~\ref{tab:dataset}, and we follow the public splits for each of them. 

\begin{table}[!t]
\caption{Details of dataset statistics.}
\centering
\label{tab:dataset}
\resizebox{\columnwidth}{!}{
\renewcommand{\arraystretch}{1}
\begin{tabular}{c c c c c c c}
\toprule
Datasets & \#Nodes & \#Edges & \#Classes & \#Features & Sparsity & Homophily \\

\midrule
Citeseer & 3,327 & 4,732 & 6 & 3,703 & 0.09\% & 0.74 \\

Cora & 2,708 & 5,429 & 7 & 1,433 & 0.15\% & 0.81 \\

Ogbn-arxiv & 169,343 & 1,166,243 & 40 & 128 & 0.01\% & 0.65 \\

\midrule

Flickr & 89,250 & 899,756 & 7 & 500 & 0.02\% & 0.33 \\

Reddit & 232,965 & 57,307,946 & 41 & 602 & 0.09\% & 0.78 \\

\bottomrule
\end{tabular}
}
\end{table}

\begin{table*}[!t]
\caption{Overall node classification accuracy on the test split of datasets. For Citeseer, Cora, and Ogbn-arxiv, we report their transductive performance. For Flickr and Reddit, we report their inductive performance. \textit{Full} indicates the performance with the original graph. The best results are highlighted in bold and grey. and the second-best results are \underline{underlined}.
Note that * indicates that we re-implement these methods to guarantee the comparison is fair since they have different GNN models for condensation in the original paper and thus the results are different.
}
\centering
\label{tab:overall}
\resizebox{0.98\linewidth}{!}{
\begin{tabular}{c c|c c c c|c c c c c c|c}
\toprule
Dataset & ratio (\%)  & Random & Herding & K-Center & Coarsening & DCG & GCond & SFGC* & SGDD* & MTT & GCSR & Full \\
\midrule
\midrule
\multirow{3}{*}{Citeseer} 
&0.9 &54.4$\pm$4.4  &57.1$\pm$1.5  &52.4$\pm$2.8  &52.2$\pm$0.4  &66.8$\pm$1.5 &\underline{70.5$\pm$1.2} &66.3$\pm$2.4 &\cellcolor{lgray}{\textbf{71.5$\pm$0.9}} &66.1$\pm$3.0 &70.2$\pm$1.1 &\multirow{3}{*}{71.7$\pm$0.4} \\

&1.8  &64.2$\pm$1.7  &66.7$\pm$1.0  &64.3$\pm$1.0  &59.0$\pm$0.5  &66.9$\pm$0.9 &70.6$\pm$0.9 &69.0$\pm$1.1 &\underline{71.2$\pm$0.7} &69.2$\pm$1.2 &\cellcolor{lgray}{\textbf{71.7$\pm$0.9}} \\

&3.6  &69.1$\pm$0.1  &69.0$\pm$0.1  &69.1$\pm$0.1  &65.3$\pm$0.5  &66.3$\pm$1.5 &69.8$\pm$1.4 &70.8$\pm$0.4 &70.9$\pm$1.2 &\underline{71.0$\pm$0.6} &\cellcolor{lgray}{\textbf{74.0$\pm$0.4}} \\
\midrule
\midrule
\multirow{3}{*}{Cora} 
&1.3  &63.6$\pm$3.7  &67.0$\pm$1.3  &64.0$\pm$2.3  &31.2$\pm$0.2  &67.3$\pm$1.9 &\underline{79.8$\pm$1.3} &{77.7$\pm$1.8} &{79.1$\pm$1.3} &{78.4$\pm$1.4} &\cellcolor{lgray}{\textbf{79.9$\pm$0.7}} &\multirow{3}{*}{81.4$\pm$0.6} \\

&2.6  &72.8$\pm$1.1  &73.4$\pm$1.0  &73.2$\pm$1.2  &65.2$\pm$0.6  &67.6$\pm$3.5 &\underline{80.1$\pm$0.6} &{79.3$\pm$0.8} &79.0$\pm$1.9 &79.7$\pm$0.9 &\cellcolor{lgray}{\textbf{80.6$\pm$0.8}} \\

&5.2  &76.8$\pm$0.1  &76.8$\pm$0.1  &76.7$\pm$0.1  &70.6$\pm$0.1  &67.7$\pm$2.2 &79.3$\pm$0.3 &79.4$\pm$0.5 &80.2$\pm$0.8 &\underline{80.5$\pm$0.6} &\cellcolor{lgray}{\textbf{81.2$\pm$0.9}} \\
\midrule
\midrule
\multirow{3}{*}{Ogbn-arxiv} 
&0.05 &47.1$\pm$3.9  &52.4$\pm$1.8  &47.2$\pm$3.0  &35.4$\pm$0.3  &58.6$\pm$0.4 &59.2$\pm$1.1 &59.0$\pm$1.8 &\underline{59.6$\pm$0.5} &58.7$\pm$1.7 &\cellcolor{lgray}{\textbf{60.6$\pm$1.1}} &\multirow{3}{*}{71.3$\pm$0.1} \\

&0.25 &57.3$\pm$1.1  &58.6$\pm$1.2  &56.8$\pm$0.8  &43.5$\pm$0.2  &59.9$\pm$0.3 &63.2$\pm$0.3 &\underline{64.6$\pm$0.3} &61.7$\pm$0.3 &64.2$\pm$0.5 &\cellcolor{lgray}{\textbf{65.4$\pm$0.8}} \\

&0.5 &60.0$\pm$0.9  &60.4$\pm$0.8  &60.3$\pm$0.4  &50.4$\pm$0.1  &59.5$\pm$0.3 &64.0$\pm$0.4 &\underline{65.2$\pm$0.8} &58.7$\pm$0.6 &65.1$\pm$0.7 &\cellcolor{lgray}{\textbf{65.9$\pm$0.6}} \\
\midrule
\midrule
\multirow{3}{*}{Flickr} 
&0.1 &41.8$\pm$2.0  &42.5$\pm$1.8  &42.0$\pm$0.7  &41.9$\pm$0.2  &46.3$\pm$0.2 &\underline{46.5$\pm$0.4} &{45.5$\pm$0.8} &46.1$\pm$0.3 &{45.4$\pm$0.4} &\cellcolor{lgray}{\textbf{{46.6$\pm$0.3}}} &\multirow{3}{*}{47.1$\pm$0.1} \\

&0.5 &44.0$\pm$0.4  &43.9$\pm$0.9  &43.2$\pm$0.1  &44.5$\pm$0.1  &45.9$\pm$0.1 &\cellcolor{lgray}{\textbf{47.1$\pm$0.1}} &46.0$\pm$0.4 &45.9$\pm$0.4 &46.0$\pm$0.4 &\underline{46.6$\pm$0.2} \\

&1 &44.6$\pm$0.2  &44.4$\pm$0.6  &44.1$\pm$0.4  &44.6$\pm$0.1  &45.8$\pm$0.1 &\cellcolor{lgray}{\textbf{47.1$\pm$0.1}} &46.1$\pm$0.3 &46.4$\pm$0.2 &46.2$\pm$0.4 &\underline{46.8$\pm$0.2} \\
\midrule
\midrule
\multirow{3}{*}{Reddit} 
&0.05 &46.1$\pm$4.4  &53.1$\pm$2.5  &46.6$\pm$2.3  &40.9$\pm$0.5  &\underline{88.2$\pm$0.2} &88.0$\pm$1.8 &80.0$\pm$3.1 &84.2$\pm$0.7 &81.6$\pm$1.8 &\cellcolor{lgray}{\textbf{90.5$\pm$0.2}} &\multirow{3}{*}{94.1$\pm$0.0} \\

&0.1 &58.0$\pm$2.2  &62.7$\pm$1.0  &53.0$\pm$3.3  &42.8$\pm$0.8  &89.5$\pm$0.1 &\underline{89.6$\pm$0.7} &84.6$\pm$1.6 &80.6$\pm$0.4 &85.2$\pm$1.3 &\cellcolor{lgray}{\textbf{91.2$\pm$0.2}} \\

&0.2 &66.3$\pm$1.9  &71.0$\pm$1.6  &58.5$\pm$2.1  &47.4$\pm$0.9  &\underline{90.5$\pm$1.2} &90.1$\pm$0.5 &87.9$\pm$1.2 &84.1$\pm$0.3 &87.7$\pm$1.1 &\cellcolor{lgray}{\textbf{92.2$\pm$0.1}} \\

\bottomrule
\end{tabular}
}
\end{table*}

\vspace{2pt}\noindent \textit{\textbf{Settings:}} We condense each dataset to three different condensation ratios ($r$), which stands for the ratio of synthetic node number $rN(0<r<1)$ to original node number $N$. 
In line with GCond~\cite{jin2021graph}, the condensation process involves two stages: $(1)$ a learning stage, where a 2-layer SGC with 256 hidden units is used to generate synthetic graphs, $(2)$ a test stage, where a 2-layer GCN with 256 hidden units is trained on the obtained synthetic graph from the first stage, and then the trained model is tested on the original test set.
For each setting, we generate 5 synthetic graphs, test each synthetic graph 10 times, and report the average node classification accuracy with standard deviation. 

\vspace{2pt}\noindent \textit{\textbf{Hyper-parameters:}} 
At the learning stage, We perform a grid search to select hyper-parameters on the following searching space: synthetic training steps $N$ is tuned amongst $\{5, 20, 50, 100\}$; expert steps $M$ is searched in $\{1, 2\}$; 
the learning rate of Adam optimizer for updating synthetic node features $\eta_2$ is selected from \{0.00001, 0.0001, 0.001, 0.01\}; 
the regularization coefficient $\alpha$ and $\beta$ is chosen from the log space between $0.1$ to $1000$;
the update rate $\tau$ is tuned from $0.9$ to $1$. 
The update rate $\gamma$ is fixed to $0.5$.
The learning rate for updating synthetic networks $\eta_1$ is fixed to $0.01$. 
At the test stage, we follow the setting of GCond~\cite{jin2021graph}. 
We fix the learning rate to $0.01$, the weight decay rate to $0.0005$, and train networks for 600 epochs. 

\vspace{2pt}\noindent \textit{\textbf{Baselines:}} We compare our framework with the following baselines: 
graph coreset methods (\textit{Random}, \textit{Herding}~\cite{welling2009herding}, and \textit{K-Center}~\cite{farahani2009facility, sener2017active}), 
graph coarsening method~\cite{huang2021scaling}, 
graph-based variant of dataset condensation method (DCG~\cite{zhao2020dataset}, MTT~\cite{cazenavette2022dataset}),
and graph condensation methods (GCond~\cite{jin2021graph}, SFGC~\cite{zheng2023structure}, and SGDD~\cite{yang2023does}).

\subsection{Overall Performance}

For dataset condensation baselines, all the synthetic graphs are generated by SGC. 
We reuse most of the results in~\cite{jin2021graph} since the experiment setting is exactly the same.
We re-implement SFGC and SGDD to guarantee a fair comparison since they use GCN to generate the condensed graph dataset.
The comparison of all the baselines is shown in Table~\ref{tab:overall}.
From the table, we can observe that: 
(1) overall, GCSR achieves superior performance compared to the baselines. 
The condensed data is comparable to the original dataset for training a GNN for node classification.
(2) Compared to baselines, GCSR exhibits an improvement in accuracy with an increase in the condensed graph size, which contradicts the observations reported in prior studies and adheres to the intuition that more data leads to better performance.
This further validates that our method is effective and successfully solves the graph dataset condensation task.
(3) Moreover, compared with baselines that solely update synthetic node features such as DCG, SFGC, and MTT, our approach outperforms them by a large margin.
We attribute the improved performance to the effectively generated adjacency matrix that captures both the self-expressiveness property of node features and the information of the original graph structure.

\begin{figure*}[!t]
    \centering
  \includegraphics[width=2.0\columnwidth]{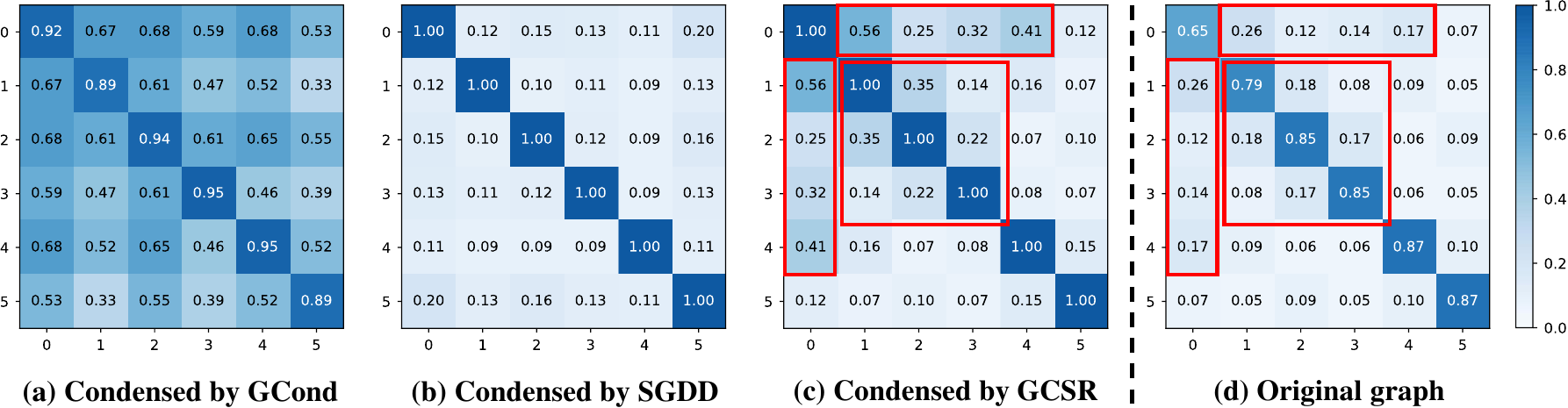}
  \caption{Cross-Class Neighborhood Similarity (CCNS) of Citeseer generated from the synthetic graph with a 3.6\% condensation ratio from (a) GCond, (b) SGDD, and (c) GCSR, as well as (d) the original graph. The axes represent the classes.}
    \label{fig:ccns}
\end{figure*}

\subsection{Cross Architecture Performance}
Consistent with state-of-the-art benchmarks~\cite{jin2021graph, yang2023does, zheng2023structure}, we evaluate the generalizability of the graph condensation framework.

\vspace{2pt}\noindent \textit{\textbf{Different GNN for Condensation:}} Here, We choose APPNP~\cite{gasteiger2018predict}, Cheby~\cite{defferrard2016convolutional}, GCN~\cite{kipf2016semi}, GraphSAGE~\cite{hamilton2017inductive}, and SGC~\cite{wu2019simplifying} to serve as the models used in the condensation phase and test the condensed dataset with GCN. The experiment is conducted on Cora with a ratio of 2.6\%.
The performance is reported in Table~\ref{tab:diffarch}.
The results reveal that:
(1) our method is model-insensitive and performs smoothly across different GNN architectures for condensation.
(2) Among all these architectures, SGC is the simplest one but surpasses others in all methods. This indicates that a more complex GNN model in condensation does not necessarily bring better optimization results since it complicates the parameter matching process.

\begin{table}[!t]
\caption{The accuracy (\%) of different GNNs used for condensation and the condensed dataset tested with GCN. The experiment is conducted on Cora with a 2.6\% ratio.}
\centering
\label{tab:diffarch}
\resizebox{1.0\linewidth}{!}{
\renewcommand{\arraystretch}{1.1}
\begin{tabular}{c |c c c c c}
\toprule
\multirow{2}{*}{Methods} &\multicolumn{5}{c}{Models} \\
\Xcline{2-6}{0.5pt}
& APPNP & Cheby & GCN & SAGE & SGC  \\
\midrule
GCOND & 73.5$\pm$2.4 & 76.8$\pm$2.1 & 70.6$\pm$3.7 & 77.0$\pm$0.7 & 80.1$\pm$0.6 \\

SFGC &78.1$\pm$0.8 &77.1$\pm$0.8 &79.1$\pm$0.7 &78.5$\pm$0.8 & 79.3$\pm$0.8 \\

SGDD &74.6$\pm$1.6 &75.9$\pm$0.2 & 73.3$\pm$0.2 &74.3$\pm$1.8 & 79.0$\pm$1.9 \\

GCSR & \cellcolor{lgray}{\textbf{80.3$\pm$1.0}} &\cellcolor{lgray}{\textbf{80.2$\pm$0.9}} & \cellcolor{lgray}{\textbf{80.1$\pm$0.9}} &\cellcolor{lgray}{\textbf{80.2$\pm$1.1}} & \cellcolor{lgray}{\textbf{80.6$\pm$0.8}} \\

\bottomrule
\end{tabular}
}
\vspace{-.3cm}
\end{table}

\begin{table}[!t]

\caption{The performance of SGC used for condensation and the condensed dataset tested with different GNNs. The average accuracy (\%) is reported. Avg. stands for the average accuracy of models except for MLP.}
\centering
\label{tab:crossarchitecture}
\resizebox{1\linewidth}{!}{
\renewcommand{\arraystretch}{1.1}
\begin{tabular}{c c| c c c c c c|c}
\toprule
\multirow{2}*{Dataset} & \multirow{2}*{Methods} & \multicolumn{6}{c|}{Models}& \multirow{2}*{Avg.} \\
\Xcline{3-8}{0.5pt}
&&MLP & APPNP & Cheby & GCN & SAGE & SGC& \\
\midrule
\midrule
\multirow{4}{*}{\makecell{Citeseer\\r=1.8\%}} 
&GCOND &58.3 &69.6 &68.3 &70.5 &66.2 &70.3 &69.0 \\

&SFGC &62.4 &69.1 &68.2 &69.0 &69.0 &69.3 &68.9 \\

&SGDD &64.9 &71.2 &70.0 &71.2 &71.4 &71.3 &71.0 \\

&GCSR &\cellcolor{lgray}{\textbf{65.7}} &\cellcolor{lgray}{\textbf{71.6}}  &\cellcolor{lgray}{\textbf{70.9}} &\cellcolor{lgray}{\textbf{71.7}} &\cellcolor{lgray}{\textbf{71.6}} &\cellcolor{lgray}{\textbf{71.8}} &\cellcolor{lgray}{\textbf{71.5}} \\

\midrule
\midrule
\multirow{4}{*}{\makecell{Cora\\r=2.6\%}} 
&GCOND &61.6 &78.5 &76.0 &80.1 &78.2 &79.3 &78.4 \\

&SFGC &65.4 &79.2 &76.4 &79.3 &79.4 &79.3 &78.7 \\

&SGDD &65.5 &77.6 &77.8 &79.0 &79.2 &77.9 &78.3 \\

&GCSR &\cellcolor{lgray}{\textbf{70.1}} &\cellcolor{lgray}{\textbf{80.6}} &\cellcolor{lgray}{\textbf{78.3}} &\cellcolor{lgray}{\textbf{80.6}} &\cellcolor{lgray}{\textbf{80.6}} &\cellcolor{lgray}{\textbf{80.8}} &\cellcolor{lgray}{\textbf{80.2}} \\
\midrule
\midrule
\multirow{4}{*}{\makecell{Ogbn-arxiv\\r=0.25\%}}
&GCOND &\cellcolor{lgray}{\textbf{45.8}} &63.4 &54.9 &63.2 &62.6 &63.7 &61.6 \\

&SFGC &45.3 &63.9 &\cellcolor{lgray}{\textbf{59.2}} &64.6 &64.7 &64.7 &63.4 \\
&SGDD &41.8 &59.8 &51.4 &61.7 &61.1 &61.1 &59.0 \\

&GCSR &44.7 &\cellcolor{lgray}{\textbf{64.4}} &58.9 &\cellcolor{lgray}{\textbf{65.4}} &\cellcolor{lgray}{\textbf{65.4}} &\cellcolor{lgray}{\textbf{65.6}} &\cellcolor{lgray}{\textbf{63.9}} \\
\midrule
\midrule
\multirow{4}{*}{\makecell{Flickr\\r=0.5\%}}
&GCOND &44.8 &45.9 &42.8 &\cellcolor{lgray}{\textbf{47.1}} &46.2 &46.1 &45.6 \\

&SFGC &44.0 &46.1 &43.6 &46.0 &46.0 &46.1 &45.6 \\
&SGDD &43.0 &45.3 &41.7 &45.9 &45.8 &45.7 &44.9 \\

&GCSR &\cellcolor{lgray}{\textbf{45.2}} &\cellcolor{lgray}{\textbf{46.3}} &\cellcolor{lgray}{\textbf{44.9}} &46.6 &\cellcolor{lgray}{\textbf{46.6}} &\cellcolor{lgray}{\textbf{46.3}} &\cellcolor{lgray}{\textbf{46.1}} \\
\midrule
\midrule
\multirow{4}{*}{\makecell{Reddit\\r=0.1\%}} 
&GCOND &42.5 &87.8 &75.5 &89.4 &89.1 &89.6 &86.3 \\

&SFGC &46.3 &84.3 &68.4 &84.6 &84.6 &86.6 &81.7 \\

&SGDD &41.6 &80.7 &69.7 &80.6 &81.1 &67.8 &76.0 \\

&GCSR &\cellcolor{lgray}{\textbf{49.5}} &\cellcolor{lgray}{\textbf{88.9}} &\cellcolor{lgray}{\textbf{80.4}} &\cellcolor{lgray}{\textbf{91.2}} &\cellcolor{lgray}{\textbf{91.0}} &\cellcolor{lgray}{\textbf{91.0}} &\cellcolor{lgray}{\textbf{88.5}} \\

\bottomrule
\end{tabular}
}

\end{table}

\vspace{2pt}\noindent \textit{\textbf{Different GNN for Test:}} We optimize synthetic graphs with SGC and test their performance under different neural architectures, i.e., MLP, APPNP, Cheby, GCN, GraphSAGE, and SGC. 
As shown in Table~\ref{tab:crossarchitecture}, our method obtains the best performance under most neural architectures and shows the best average performance. It has been well studied in~\cite{nt2019revisiting, zhu2021interpreting, ma2021unified} that different GNN models show similar low-pass filtering behaviors and the graph topology structure is a low-pass filter that provides a means to denoise the data. Our method fully retains the original structure information which contributes to its excellent transferability.

\subsection{Learned Graph Structure Visualization}

\subsubsection{Similarity of Synthetic Structure and Original Structure.} As shown in Fig.~\ref{fig:ccns}, we report the cross-class neighborhood similarity (CCNS)~\cite{ma2021homophily} of Citesser generated from the synthetic graph of GCond, SGDD, and GCSR, as well as the original graph. We do not involve SFGC as the synthetic graph learned by it is structure-free. The value of CCNS $s(c, c')$ measures the neighborhood patterns similarity of nodes in class $c$ and nodes in class $c'$ from a neighborhood label distribution perspective. It is a good metric that reflects the connection relationships between nodes in a graph. From Fig.~\ref{fig:ccns}, we can find that: 
(1) graphs learned from GCond exhibit inter-class similarity with no clear distinction because their structure is directly generated from synthetic node features via MLP. 
(2) SGDD tends to emphasize the intra-class similarity while ignoring the inter-class similarity. Likewise, it also exhibits no clear distinction in inter-class similarity. 
(3) Owing to the ability to leverage the original graph structure and the inter-node correlations within the synthetic graph, our results mimic the node connection characteristics of the original graph excellently, which leads to better performance ultimately.

\subsubsection{Interpretability of Self-expressive Reconstruction.} 
\begin{figure}[!t]
    \centering
  \includegraphics[width=1.0\columnwidth]{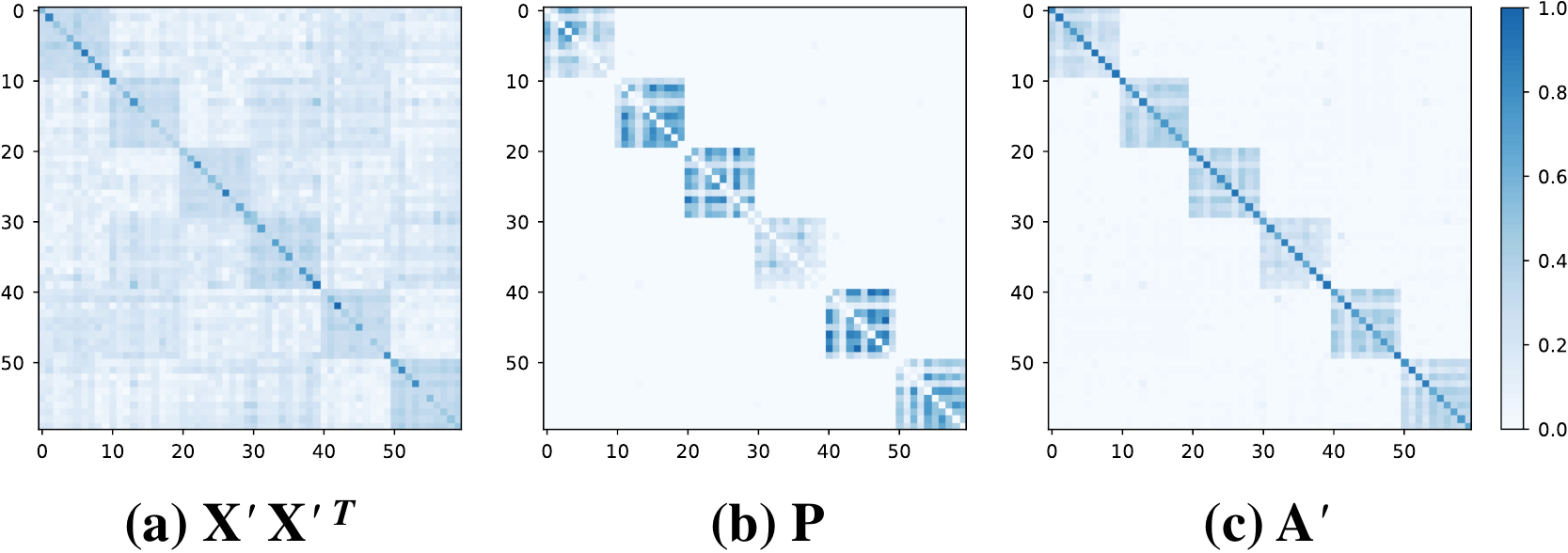}
  \caption{Visualizations of (a) the inner product of the synthetic nodes, (b) the updated probabilistic graph, and (c) the synthetic topology structure. 
  The dataset illustrated is Citeseer with the condensation ratio set to 1.8\%.}
    \label{fig:adj}
\end{figure}

We visualize the learned synthetic adjacency matrix of Citeseer as well as the components used to generate it (i.e., $\textbf{X}'\textbf{X}'^T$ and $\textbf{P}$). From Fig.~\ref{fig:adj}, we can observe that the derived synthetic adjacency matrix tends to maintain the joint connection properties of $\textbf{X}'\textbf{X}'^T$ and $\textbf{P}$. Connections between nodes of the same class tend to be retained, while connections between nodes of different classes tend to be eliminated. Such a phenomenon aligns with the connection characteristics of the homophilic graph like Citeseer, where nodes are prone to connect with similar others.

\subsection{Learned Node Feature Visualization} 
\begin{figure}[!t]
    \centering
  \includegraphics[width=1.0\columnwidth]{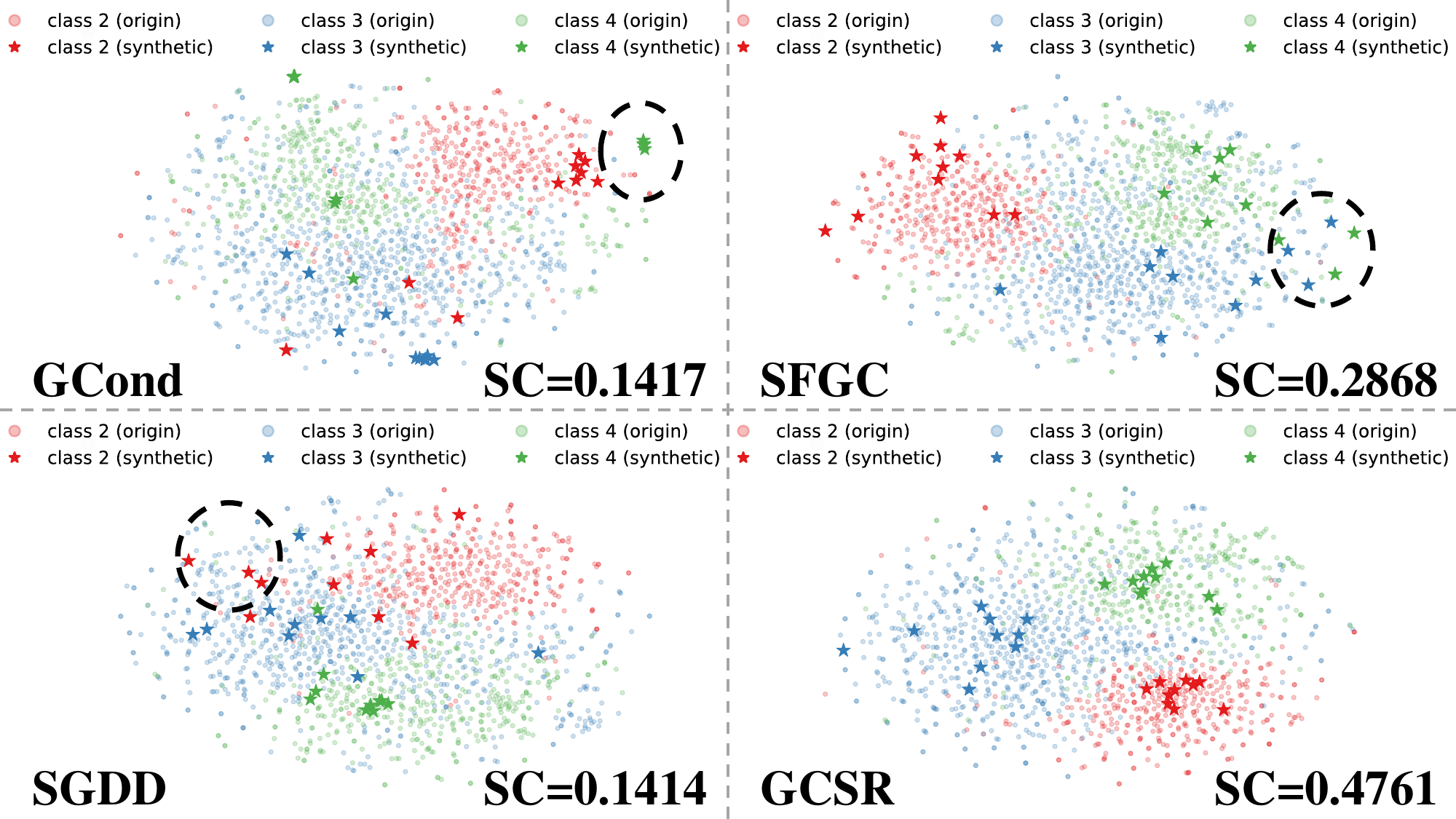}
  
  \caption{Distributions of the synthetic nodes condensed by four methods (GCond, SFGC, SGDD, and GCSR) on Cora under a 2.6\% condensation ratio. $\text{SC}$ represents the average silhouette coefficient of the synthetic data.}
    \label{fig:tsne}
\end{figure}

To evaluate whether our method can capture more information from the original graph, we use t-SNE~\cite{van2008visualizing} to visualize the real and synthetic node features of Cora condensed by GCond, SFGC, SGDD, and GCSR.
For a clearer display, we only show nodes from 3 out of 7 classes.
We also calculate the average silhouette coefficient~\cite{rousseeuw1987silhouettes} 
(a metric that measures the distance between samples of different classes and a higher coefficient is better)
of synthetic data generated from each method.
Fig.~\ref{fig:tsne} shows that:
(1) data learned by GCond, SFGC, and SGDD is not well conformed to the distribution of the original data, and some nodes are mixed up with nodes of other classes.
(2) Data learned by GCSR obeys the distribution of the original data and exhibits a distinct separation between different classes. This could lead to better classification performance and evident distribution characteristics. 
(3) GCSR achieves the highest silhouette coefficient, indicating the learned node features of different classes are easy to separate.

\begin{figure}[!t]
    \centering
  \includegraphics[width=0.99\linewidth]{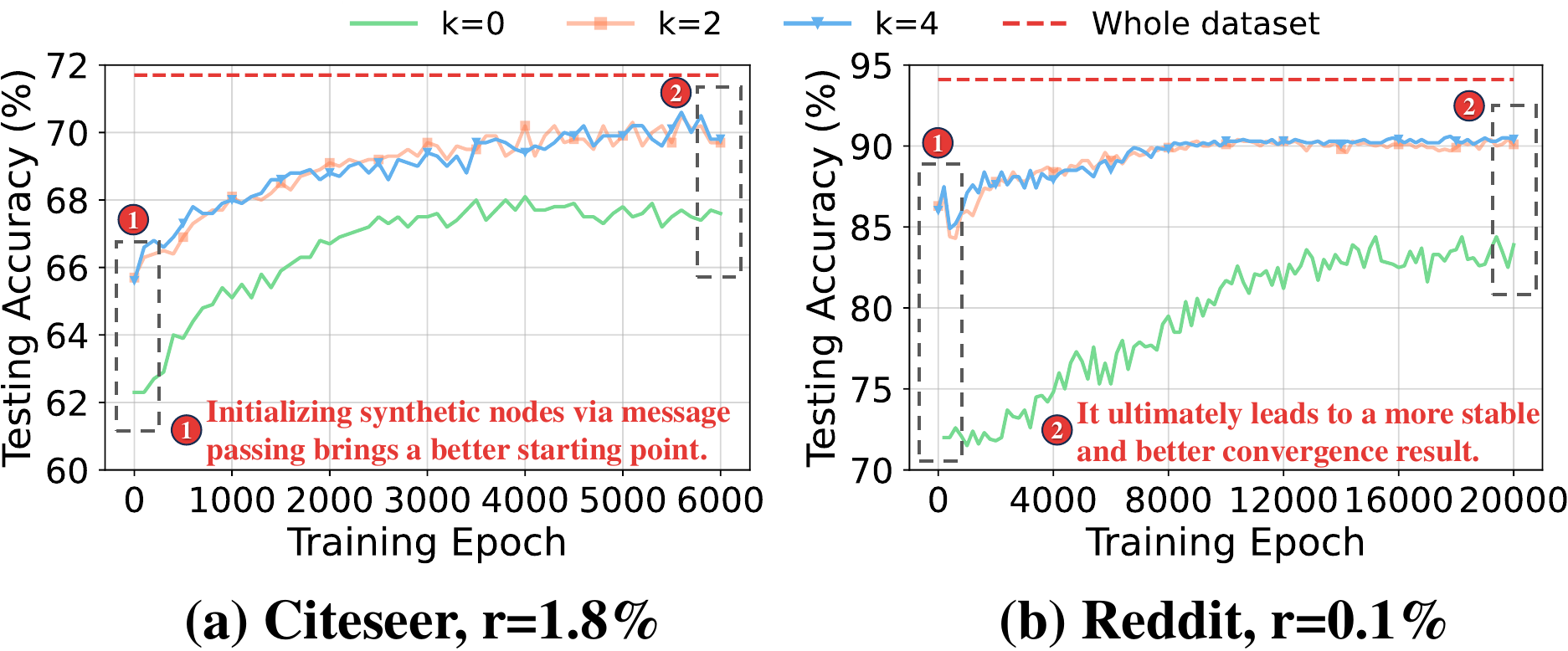}
  
  \caption{Training curve of different initialization methods. $k$ stands for the hop of neighborhood being captured. $k=0$ will have no message passing and have the same performance as a random selection from original nodes.}
    \label{fig:mp}
\end{figure}

\subsection{Node Initialization} 

To evaluate the proposed message passing initialization for node features, we conduct extra experiments with one transductive dataset (Citeseer) and one inductive dataset (Reddit) under different initialization methods. 
To remove interference from other modules, we do not reconstruct the self-expressive adjacency matrix, i.e., we use the identity matrix as the graph structure in the synthetic dataset. 
Fig.~\ref{fig:mp} illustrates the node classification training curve of different initialization.
We can make the following observations: 
(1) Initializing synthetic nodes via message passing brings a better starting point. 
We attribute such improvement to the fusion of the original graph structure and node features that enhance the synthetic data and the low-pass filtering properties of message passing that denoise the data. 
(2) Message passing initialization ultimately leads to a more stable and better convergence result.
Generally, the end performance of Citeseer is about 1.3\% higher and that of Reddit is about 6.1\% higher. 
Such improvements are marvelous since there is no extra cost. 
(3) The results are not significantly influenced by the number of hops of the neighborhood.
Empirically, setting $k$ to 2 or 3 is a more robust and general choice.

\subsection{Ablation Study}

In GCSR, two components help broadcast the original graph structure to the synthetic graph, i.e., message passing initialization and self-expressive reconstruction. 
We conduct an ablation study to analyze the contribution of each component. 
As seen in Table~\ref{tab:ablation}, the joint use of two components brings about state-of-the-art performance. 
The results support our claim that both components contribute to leveraging information from the original graph. 
Message passing initialization transfers the implicit information of the original graph structure to the synthetic node features.
Self-expressive reconstruction generates the interpretable graph structure that combines the information of the correlation between the node embeddings and the global structure information from the original graph.
It is worth mentioning that, as a novel approach to initialize synthetic nodes, message passing initialization is independent of other modules and can significantly improve the condensation results even without adjacency matrices.

\begin{table}[!t]
\caption{Evaluation of the effectiveness of each module. \textit{MPI} denotes message passing initialization for node featuers and \textit{SER} represents self-expressive reconstruction. The condensation model is SGC. The performance is shown in test accuracy (\%) of GCN.}
\centering
\label{tab:ablation}
\resizebox{\columnwidth}{!}{
\renewcommand{\arraystretch}{1.2} % default is 1.0
\begin{tabular}{c c |c c c c c}
\toprule
\multirow{2}{*}{MPI} & \multirow{2}{*}{SER} & Citeseer & Cora & Ogbn-arxiv & Flickr & Reddit \\
& & r=1.8\% & r=2.6\% & r=0.25\% & r=0.5\% & r=0.1\% \\
\midrule
- & - & 69.2$\pm$1.2 & 79.7$\pm$0.9 & 64.2$\pm$0.5 & 46.0$\pm$0.4 & 85.2$\pm$1.3 \\

\checkmark & - & 70.1$\pm$0.8 & 80.4$\pm$0.8 & 64.2$\pm$1.1 & 46.4$\pm$0.4 & 90.4$\pm$0.7 \\

\checkmark & \checkmark & \cellcolor{lgray}{\textbf{71.7$\pm$0.9}} & \cellcolor{lgray}{\textbf{80.6$\pm$0.8}} & \cellcolor{lgray}{\textbf{65.4$\pm$0.8}} & \cellcolor{lgray}{\textbf{46.6$\pm$0.2}} & \cellcolor{lgray}{\textbf{91.2$\pm$0.2}}  \\

\bottomrule
\end{tabular}
}
\end{table}

\section{Conclusion}

In this paper, we present a novel framework \textbf{GCSR} which learns synthetic graphs via self-expressive graph structure reconstruction.
It fills the gap in existing methods by making full use of the original graph structure and the correlations between node features to generate an interpretable synthetic graph structure.
Extensive experimental results demonstrate the superiority of our proposed method.
In the future, we plan to explore a more comprehensive reconstruction strategy that broadcasts more information from the original graph to the synthetic graph (e.g., sparsity and heterogeneity) to enhance the generalizability and robustness of synthetic graphs.
\section*{Acknowledgement}
This work was sponsored by National Key Research and Development Program of China under Grant No.2022YFB3904204, National Natural Science Foundation of China under Grant No. 62102246, 62272301, and Provincial Key Research and Development Program of Zhejiang under Grant No. 2021C01034. Part of the work was done when the students were doing internships at Yunqi Academy of Engineering.

%\clearpage
\bibliographystyle{ACM-Reference-Format}
\balance
\bibliography{sample-base}
\clearpage
\appendix

\section{The Effectiveness of Message Passing Initialization}

\begin{table}[!t]
\caption{The effectiveness of MPI.}
\centering
\label{tab:MPI_ef}
\resizebox{\columnwidth}{!}{
\renewcommand{\arraystretch}{1}
\begin{tabular}{c|c c c c c}
\toprule
Dataset & GCSR & SFGC & SFGC+MPI & GCond & 	GCond+MPI\\
\midrule
Citeseer (r=1.8\%) & \cellcolor{lgray}{\textbf{74.0 $\pm$ 0.4}} & 70.8 $\pm$ 0.4 & 71.8 $\pm$ 0.4 $\uparrow$ & 69.8 $\pm$ 1.4 & 71.3 $\pm$ 0.9 $\uparrow$\\
Reddit (r=0.1\%) & \cellcolor{lgray}{\textbf{91.2 $\pm$ 0.2}} & 84.6 $\pm$ 1.6 & 90.5 $\pm$ 0.2 $\uparrow$ & 89.6 $\pm$ 0.7 & 90.2 $\pm$ 0.2 $\uparrow$\\

\bottomrule
\end{tabular}
}
\end{table}

\begin{table}[!t]
\caption{The sensitivity analysis of $k$ in MPI.}
\centering
\label{tab:MPI_k}
\resizebox{\columnwidth}{!}{
\renewcommand{\arraystretch}{1}
\begin{tabular}{c|c c c c c c}
\toprule
Dataset & 0 & 2 & 4 & 6 & 8 & 10\\
\midrule
Citeseer(r=1.8\%) & 69.2 $\pm$ 1.2 & 70.2 $\pm$ 0.8	 & 70.2 $\pm$ 0.8 & 70.3 $\pm$ 0.7 & 70.4 $\pm$ 0.7 & 70.3 $\pm$ 0.7\\

\bottomrule
\end{tabular}
}
\end{table}

\begin{table}[!t]
\caption{The effectiveness of graph regularizer initialization of GCSR.}
\centering
\label{tab:reg_ef}
\resizebox{\columnwidth}{!}{
\renewcommand{\arraystretch}{1}
\begin{tabular}{c|c c}
\toprule
 Method& Citeseer (r=1.8\%) & Reddit (r=0.1\%)\\
\midrule
GCSR & \cellcolor{lgray}{\textbf{71.8 $\pm$ 0.9}} & \cellcolor{lgray}{\textbf{91.2 $\pm$ 0.2}} \\
Sub-structure Init & 69.5 $\pm$ 2.0 & 91.0 $\pm$ 0.3 \\
Random Init & 64.8 $\pm$ 2.6 & 72.2 $\pm$ 2.5\\

\bottomrule
\end{tabular}
}
\end{table}

\begin{table}[!t]
\caption{The sensitive analysis of regularizer update parameter $\tau$.}
\centering
\label{tab:reg_tau}
\resizebox{\columnwidth}{!}{
\renewcommand{\arraystretch}{1}
\begin{tabular}{c|c c c c c}
\toprule
Dataset & 0.5 & 0.9 & 0.99 & 0.999 & 1 \\
\midrule
Citeseer(r=1.8\%) & 70.2 $\pm$ 1.0 & \cellcolor{lgray}{\textbf{71.8 $\pm$ 0.9}}	 & 70.2 $\pm$ 1.2 & 68.7 $\pm$ 1.7 & 68.1 $\pm$ 1.8 \\
\bottomrule
\end{tabular}
}
\end{table}

\begin{table}[!t]
\caption{The sensitive analysis of regularizer update parameter $\gamma$.}
\centering
\label{tab:reg_gamma}
\resizebox{\columnwidth}{!}{
\renewcommand{\arraystretch}{1}
\begin{tabular}{c|c c c c}
\toprule
Dataset & 0.1 & 0.5 & 0.9 & 1 \\
\midrule
Citeseer(r=1.8\%) & 71.6 $\pm$ 0.9 & \cellcolor{lgray}{\textbf{71.8 $\pm$ 0.9}}	 & 71.6 $\pm$ 0.9 & 70.6 $\pm$ 0.8\\
\bottomrule
\end{tabular}
}
\end{table}

\begin{table}[!t]
\caption{The sensitivity analysis of $\mathbf{\alpha}$.}
\centering
\label{tab:ps_alpha}
\resizebox{\columnwidth}{!}{
\renewcommand{\arraystretch}{1}
\begin{tabular}{c|c c c c c }
\toprule
Dataset & 0 & 0.1 & 0.5 & 1 & 10 \\
\midrule
Citeseer(r=1.8\%) &70.1 $\pm$ 0.8 & 70.2 $\pm$ 1.0	 & 71.1 $\pm$ 0.8 & \cellcolor{lgray}{\textbf{71.8 $\pm$ 0.9}} & 70.1 $\pm$ 1.0 \\

Ogbn-arxiv (r=0.05\%) & 59.9 $\pm$ 1.1 & \cellcolor{lgray}{\textbf{60.6 $\pm$ 1.1}}	 & 59.2 $\pm$ 1.3 & 58.5 $\pm$ 1.8 & 58.8 $\pm$ 0.7 \\
\bottomrule
\end{tabular}
}
\end{table}

\begin{table}[!t]
\caption{The sensitivity analysis of $\mathbf{\beta}$.}
\centering
\label{tab:ps_beta}
\resizebox{\columnwidth}{!}{
\renewcommand{\arraystretch}{1}
\begin{tabular}{c|c c c c c }
\toprule
Dataset & 0 & 0.1 & 0.5 & 1 & 10 \\
\midrule
Citeseer(r=1.8\%) &71.6 $\pm$ 0.9 & \cellcolor{lgray}{\textbf{71.8 $\pm$ 0.9}}	 & 71.6 $\pm$ 1.0 & 71.6 $\pm$ 0.9 & 71.3 $\pm$ 1.0 \\

Ogbn-arxiv (r=0.05\%) & 59.8 $\pm$ 1.3 & 60.3 $\pm$ 1.3	 & 60.2 $\pm$ 1.3 & \cellcolor{lgray}{\textbf{60.6 $\pm$ 1.1}} & 60.5 $\pm$ 1.0 \\
\bottomrule
\end{tabular}
}
\end{table}

GCSR introduces message-passing initialization (MPI), which is a novel approach in the field of graph condensation. To further verify the effectiveness of MPI, we implement it on two other graph condensation frameworks and check whether MPI brings performance enhancement. The result is shown in Table~\ref{tab:MPI_ef}. We could observe that with MPI, the performance of SFGC and GCond both improve significantly, which validates the effectiveness of our proposed MPI module. 
Nevertheless, our method still performs better, which illustrates the necessity of SER.

Furthermore, we evaluate the sensitivity of parameter $k$ in the MPI module. We solely implement MPI on GCSR and get the result of different $k$ on Table~\ref{tab:MPI_k}. Consistent with our paper, MPI could significantly improve condensation performance. However, the results are not significantly influenced by $k$. As illustrated in the table above, even if we set $k$ to 8 or 10, the improvement compared to $k$=2 is very small.

\section{Graph Regularizer Update Analysis}

We initialize the graph regularizer with a probabilistic adjacency matrix to avoid the inconsistency between the dimensions of adjacency matrices of the synthetic graph and the original graph.
To further verify the effectiveness of our initialization, we compare our initialization with sub-structure initialization of DosCond~\cite{jin2022condensing} and random initialization. The result is shown in Table~\ref{tab:reg_ef}. We can observe that the probabilistic adjacency matrix initialization achieves the best performance.

Furthermore, update momentum ratio $\tau$ and $\gamma$ play important roles in updating the graph regularizer in Eqs.~\ref{formula:14} and~\ref{formula:gamma}.
We evaluate their impact by setting different $\tau$ and $\gamma$ as shown in Table~\ref{tab:reg_tau} and Table~\ref{tab:reg_gamma}. 
Here $\tau=1$ means not update $\textbf{P}$ and $\gamma=1$ means not update $\textbf{Z}_h$.
We could observe that these two regularizer updates contribute to the final performance. 
Moreover, the performance is more sensitive to $\tau$, so it is recommended to search for more $\tau$ in the real application.

\section{Graph Reconstruction Analysis}

In this section, we analyze the impact of hyper-parameter $\alpha$ and $\beta$ in reconstructing the graph structure $\mathbf{Z}$ in Eq.~\ref{formula:9} on Citeseer and Ogbn-arxiv as shown in Table~\ref{tab:ps_alpha} and Table~\ref{tab:ps_beta}. 
Here, $\alpha=0$ means we only apply regularizer $\textbf{Z}_h$ and $\beta=0$ indicates we only apply regularizer $\textbf{P}$. 
We could observe that adding each regularizer could enhance the performance and the joint use of two regularizers brings the best performance. 
Moreover, the hyperparameter is recommended to search in (0,1].

\end{document}